# Minimal Change and Bounded Incremental Parsing

Mats Wirén
Fachrichtung 8.7, Computerlinguistik
Universität des Saarlandes
Postfach 1150
D-66041 Saarbrücken, Germany
wiren@coli.uni-sb.de

## Abstract

Ideally, the time that an incremental algorithm uses to process a change should be a function of the size of the change rather than, say, the size of the entire current input. Based on a formalization of "the set of things changed" by an incremental modification, this paper investigates how and to what extent it is possible to give such a guarantee for a chart-based parsing framework and discusses the general utility of a minimality notion in incremental processing.[1]

## 1 Introduction

### 1.1 Background

Natural-language computing has traditionally been understood as a "batch-mode" or "once-only" process, in which a problem instance $P$ (say, a text) is mapped as a whole to a solution $S$ (such as an analysis of the text). However, in highly interactive and real-time applications — for example, grammar checking, structure editing and on-line translation — what is required is efficient processing of a sequence of small *changes* of a text. Exhaustive recomputation is then not a feasible alternative. Rather, to avoid as much recomputation as possible, each update cycle must re-use those parts of the previous solution that are still valid. We say that an algorithm is *incremental* if it uses information from an old solution in computing the new solution.

The problem of incremental processing can be stated as follows, using a notation similar to that of Alpern et al. [1]: Assume given a problem instance $P$ (a representation of the current input), a solution $S$ (the current output), and a modification $\Delta_P$ to $P$.[2] The modification results in a new problem instance $P' = P \oplus \Delta_P$, where $\oplus$ is a composition operator. The task of an incremental algorithm is then to produce an update $\Delta_S$ in the old solution such that $S \oplus \Delta_S$ is a solution to $P \oplus \Delta_P$ (see figure 1). At this point, nothing is stipulated about the amount of information in $S$ that should be re-used in $S'$.

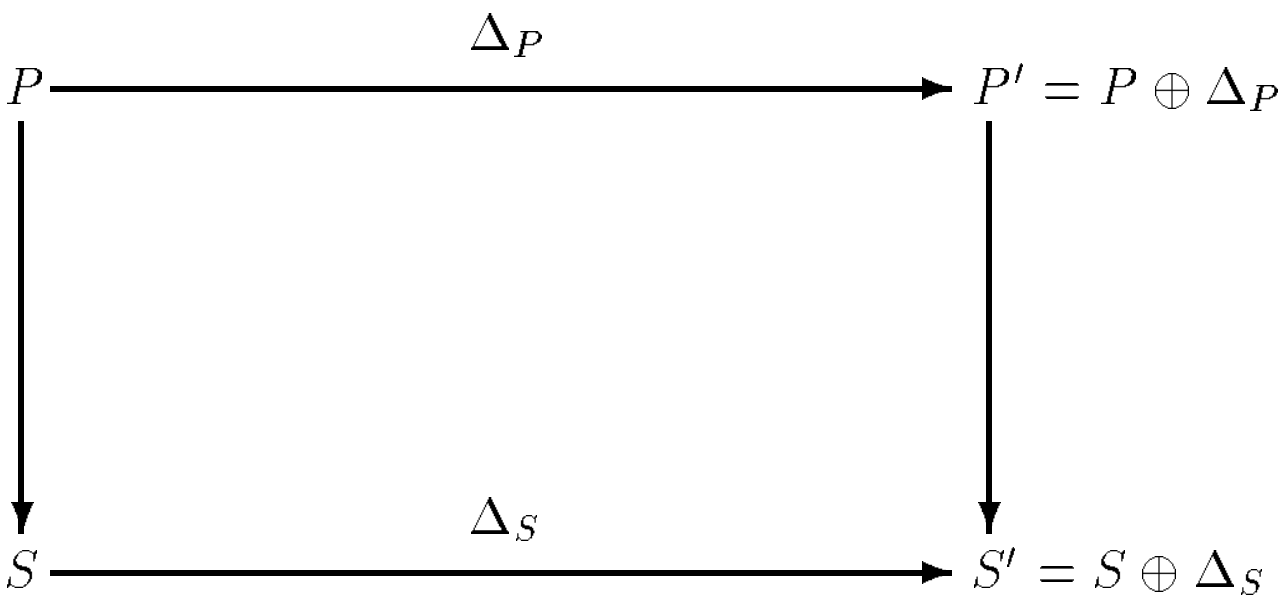

Figure 1: Batch-mode versus incremental computation.

To show properties such as correctness and complexity of incremental algorithms, it is necessary to establish a formal measure of "the set of things changed". This measure should capture the minimal change resulting from a modification and, moreover, should be independent of any particular algorithms for incremental update. One way of achieving this is to compare the results obtained by *batch-mode* processing of the inputs before and after the change, respectively (Wirén and Rönnquist [15, 17]): By forming the "difference" between the batch-mode solutions $S$ and $S'$ obtained before and after a modification $\Delta_P$ to $P$, we obtain a parameter $\Delta_{S_{min}}$ which captures the minimal change in a way which is indeed independent of the incremental update. Given that $\Delta_{S_{min}}$ corresponds precisely to what any sound and complete incremental algorithm *must* do, it can be used as a basis for correctness proofs for such algorithms (given that the batch-mode algorithm is correct).

Furthermore, $\Delta_{S_{min}}$ can be used as a basis of complexity analyses: Ideally, each update cycle of an incremental algorithm should expend an amount of work which is a polynomial function of the size of the change, rather than, say, the size of the entire current input. However, making this notion precise in a way which is independent of particular incremental algorithms is not always straightforward. Two early approaches along these lines are Goodwin [3, 4] (reason maintenance) and Reps [11] (language-based editing). More recently, Alpern et al. [1] and Ramalingam and Reps [9, 10] have provided a framework for analysing incremental algorithms, in which the basic measure used is the sum of the sizes of the changes in the input and output. This framework assumes that the modification of the input can be carried out in $O(|\Delta_P|)$ time, where the generic notation $|X|$ is used for the size of $X$. Furthermore, it assumes that $|\Delta_{S_{min}}|$ denotes the minimal $|\Delta_S|$ such that $S \oplus \Delta_S$ solves $P \oplus \Delta_P$. Alpern et al. then define

$$\delta = |\Delta_P| + |\Delta_{S_{min}}|$$

as the *intrinsic size* of a change.

[1] I would like to thank Ralph Rönnquist as well as Gregor Erbach and other colleagues in Saarbrücken for discussions on the material presented here, Peter Fritzson for originally alerting my attention to Ramalingam and Reps' paper, and the anonymous referees. This research has been funded by the German Science Foundation (DFG) through the Sonderforschungsbereich 314, project N3 (BiLD).

[2] A terminological note: we use "input change" and "modification" as well as "output change" and "update" synonymously.

The choice of $\delta$ is motivated as follows: $|\Delta_P|$, the size of the modification, is in itself too crude a measure, since a small change in problem instance may cause a large change in solution or vice versa. $|\Delta_{S_{min}}|$ is then chosen as a measure of the size of the change in the solution, since the time for updating the solution can be no less than this. The $\delta$ measure thus makes it possible to capture how well a particular algorithm performs relative to the amount of work that *must* be performed in response to a change.

An incremental algorithm is said to be *bounded* if it can process any change in time $O(f(\delta))$, that is, in time depending only on $\delta$. Intuitively, this means that it only processes the "region" where the input or output changes. Algorithms of this kind can then be classified according to their respective degrees of boundedness (see Ramalingam and Reps [10, section 5]). For example, an algorithm which is linear in $\delta$ is asymptotically optimal. Furthermore, an incremental algorithm is said to be *unbounded* if the time it takes to update the solution can be arbitrarily large for a given $\delta$.

It might seem that what has been discussed so far has little relevance to natural-language processing, where incrementality is typically understood as the *piecemeal* assembly of an analysis during a single left-to-right[3] pass through a text or a spoken utterance. In particular, incrementality is often used as a synonym for *interleaved* approaches, in which syntax and semantics work in parallel such that each word or phrase is given an interpretation immediately upon being recognized (see, for example, Mellish [7] and Haddock [5]). However, the two views are closely related: The "left-to-right view" is an idealized, psycholinguistically motivated special case, in which the only kind of change allowed is addition of new material at the end of the current input, resulting in piecemeal expansion of the analysis. Moreover, the interleaving is just a consequence of the fact that every piece of new input must, in some sense, be fully analysed in order to be integrated with the old analysis.

To distinguish this special case from the general case, in which arbitrary changes are allowed, Wirén [15] refers to them as *left-to-right* (*LR*) *incrementality* and *full incrementality*, respectively. The former case corresponds to *on-line* analysis — that each prefix of a string is parsed (interpreted) before any of the input beyond that prefix is read (Harrison [6, page 433]). The latter case has long been studied in interactive language-based programming environments (for example, Ghezzi and Mandrioli [2]), whereas the only previous such work that we are aware of in the context of natural-language processing is Wirén and Rönnquist [14, 15, 16, 17].

[3] Strictly speaking front-to-back or beginning-to-end.

### 1.2 The Problem

The aim of this paper is to begin to adapt and apply the notion of bounded incremental computation to natural-language parsing, using a method for establishing minimal change previously introduced by Wirén and Rönnquist [15, 17]. To this end, the paper shows how the $\delta$ parameter can be defined in a fully incremental, chart-based parsing framework, briefly describes a previous, unbounded algorithm, and then shows how a polynomially bounded algorithm can be obtained.

## 2 Batch-Mode Chart Parsing

An incremental problem can be defined by specifying its batch-mode version and the set of allowable modifications. We thus begin by specifying batch-mode chart parsing, restricting ourselves to a standard context-free grammar without cyclic or empty productions.

**Definition 1 (Chart)** A *chart* is a directed graph $C = \langle V, E \rangle$ such that $V$ is a finite, non-empty set of vertices and $E \subseteq V \times V \times R$ is a finite set of edges, where $R$ is the set of dotted context-free rules obtained from the grammar.[4]

The vertices $v_1, \ldots, v_{n+1} \in V$ correspond to the linear positions between the tokens $\tau = t_1 \cdots t_n$ of an $n$-token text.[5] An edge $e \in E$ between vertices $v_i$ and $v_j$ carries information about a (partially) analysed constituent between the corresponding positions.

The algorithm makes use of an agenda (see Thompson [12]). Agenda tasks are created in response to tokens being read and edges being added to the chart, and may be ordered according to their priorities. To define the agenda, we make use of the set of possible tokens *Tkns* and the set of possible edges *Edgs*.

**Definition 2 (Agenda)** We define the *agenda* as $Agda \subseteq Tkns \cup Edgs \cup (Edgs \times Edgs)$. We refer to the three types of tasks that it contains as *scanning*, *prediction* and *combination tasks*, respectively.

[4] For brevity, we omit a fourth edge component corresponding to the set of (partial) parse trees according to the grammar and lexicon (assuming that only the topmost portion of a tree corresponding to the dotted rule needs to be stored in an edge).

[5] We shall use $\tau$ interchangeably to denote a sequence and a set of tokens.

Each agenda task is executed by a step of the algorithm below. We specify two versions of batch-mode chart parsing — the basic bottom-up (strictly speaking, left-corner) and top-down (Earley-style) strategies — assuming that the one or the other is chosen.

**Algorithm 1 (Batch-mode chart parsing)**

**Input:** A sequence of tokens $\tau = t_1 \cdots t_n$.

**Output:** A chart.

**Initialization:** If the top-down strategy is used, then add an agenda task corresponding to an initial top-down prediction $\langle v_1, v_1, S \rightarrow \mathbf{.}\alpha \rangle$ for each rule $S \rightarrow \alpha$, where $S$ is the start category of the grammar.

**Method:** For each token, create a scanning task. While the agenda is not empty, remove the next task and execute the corresponding step below:

**Scan:** Given a token $t$ at position $j$, for each lexical entry of the form $X \rightarrow t$, add an edge $\langle v_j, v_{j+1}, X \rightarrow t \mathbf{.} \rangle$.[6] Add resulting new tasks to the agenda.

**Predict 1 (Bottom-up):** If the edge is of the form $\langle v_j, v_k, X \rightarrow \alpha \mathbf{.} \rangle$, then, for each rule of the form $Y \rightarrow X\gamma$, add an edge $\langle v_j, v_j, Y \rightarrow \mathbf{.} X\gamma \rangle$ unless it already exists. Add resulting new tasks to the agenda.

**Predict 2 (Top-down):** If the edge is of the form $\langle v_i, v_j, X \rightarrow \alpha \mathbf{.} Y\beta \rangle$, then, for each rule of the form $Y \rightarrow \gamma$, add an edge $\langle v_j, v_j, Y \rightarrow \mathbf{.}\gamma \rangle$ unless it already exists. Add resulting new tasks to the agenda.

**Combine:** If the first edge is of the form $\langle v_i, v_j, X \rightarrow \alpha \mathbf{.} Y\beta \rangle$ and the second is of the form $\langle v_j, v_k, Y \rightarrow \gamma \mathbf{.} \rangle$, then add an edge $\langle v_i, v_k, X \rightarrow \alpha Y \mathbf{.} \beta \rangle$. Add resulting new tasks to the agenda.

# 3 Incremental Chart Parsing

## 3.1 The Problem

The overall incremental process can be thought of as a change–update loop, where each change of the input is immediately followed by a corresponding update of the output. To completely specify the state of this process, we shall make use of a configuration consisting of (a representation of) an input text $\tau$, a chart $C$ and an edge-dependency relation $\mathcal{D}$ (to be defined in section 4). The problem of incremental chart parsing can then be specified abstractly as a mapping

$$f(\langle \tau, C, \mathcal{D} \rangle, \Delta_\tau) \mapsto \langle \tau', C', \mathcal{D}' \rangle$$

from an old configuration and a modification $\Delta_\tau$ to a new configuration. We shall allow two kinds of change, namely, insertion and deletion of $m \geq 1$ contiguous tokens. We assume that a modification $\Delta_\tau$ is given as a vertex pair $v_j, v_{j+m} \in V$ defining the update interval and, in the case of an insertion, a sequence of tokens $\tau = t_j \cdots t_{j+m}$. We furthermore assume that either the bottom-up or top-down strategy is chosen throughout a change–update session, and, in the latter case, that the top-down initialization is made before the session is started.

## 3.2 A General Vertex Mapping

How can the minimal change $\Delta_{S_{min}}$ be defined in a chart-based framework? One way of doing this is to compare the charts $C = \langle V, E \rangle$ and $C' = \langle V', E' \rangle$ that are obtained by *batch-mode* parsing of the texts before and after a change, respectively. We thereby obtain a measure which is independent of particular incremental update algorithms. Intuitively, only those edges that are in $E$ but not in $E'$ must be removed, and only those edges that are in $E'$ but not in $E$ must be generated anew. If the change is small, then a large fraction of the edges are in $E \cap E'$ (that is, are unchanged).

However, to be able to compare the edge sets in the two charts, we must first establish a one-to-one mapping between their vertices. Let us consider the case in which a single token $t_i$ is deleted from an $n$-token text. The problem is that, because of the removed token, the two vertices $v_i$ and $v_{i+1}$ would seem to correspond to a single vertex in $V'$. However, we can regard this single vertex as consisting of a "left half" and a "right half", which we assign different indices. In other words, after having increased each index of $v'_{i+1}, \ldots, v'_n \in V'$ by one, we "split" vertex $v'_i$ and assign the index $i+1$ to its "right half". The incoming non-predicted edges as well as (looping) top-down predictions at the split vertex are then associated with its left half, and the outgoing non-predicted edges as well as (looping) bottom-up predictions are associated with its right half.[7] The reason for dividing the predicted edges in this way is that a top-down prediction is made at the ending vertex of the triggering edge (that is, from the left), whereas a bottom-up prediction is made at the starting vertex of the triggering edge (that is, from the right).

The mapping can be generalized to the case in which $m$ contiguous tokens are deleted. This is done by increasing the index of each vertex from the "right half" of the split vertex and onwards by $m$ (instead of one). Furthermore, by using the same mapping but in the opposite direction, we can also cover insertion of $m$ contiguous tokens. To express this generalized mapping, assume that $\tilde{V}$ is the set of vertices of the larger chart and $V$ is that of the smaller chart. A deletion of $m$ contiguous tokens then involves a mapping from $\tilde{V}$ to $V$ and an insertion of $m$ tokens involves a mapping from $V$ to $\tilde{V}$. In terms of the indexing that holds *before* the vertices in $V$ are renumbered, and assuming that $\tilde{V}$ has $n+1$ vertices, we obtain the following bidirectional

[6] We refer to the new edge as a *lexical* edge.

[7] As mentioned above, we assume that only the one *or* the other strategy is used, so that it is known beforehand which kind of predictions the chart contains.

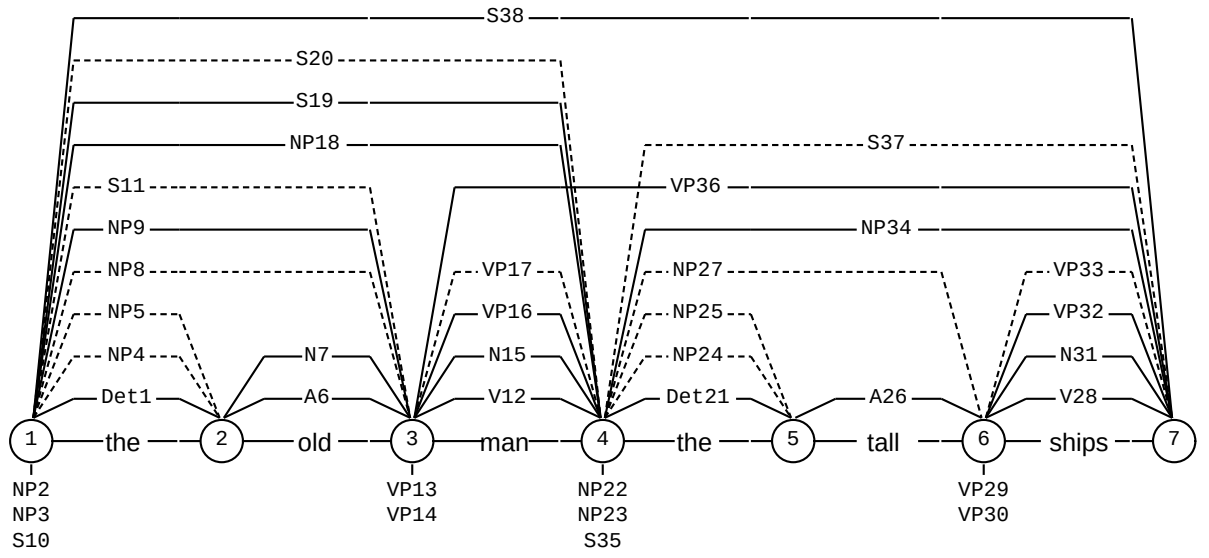

Figure 2: Chart of the sentence "The old man the tall ships" under bottom-up parsing. Inactive edges are drawn using continuous lines, active edges using dashed lines, and predicted (looping) edges are depicted below the vertices.

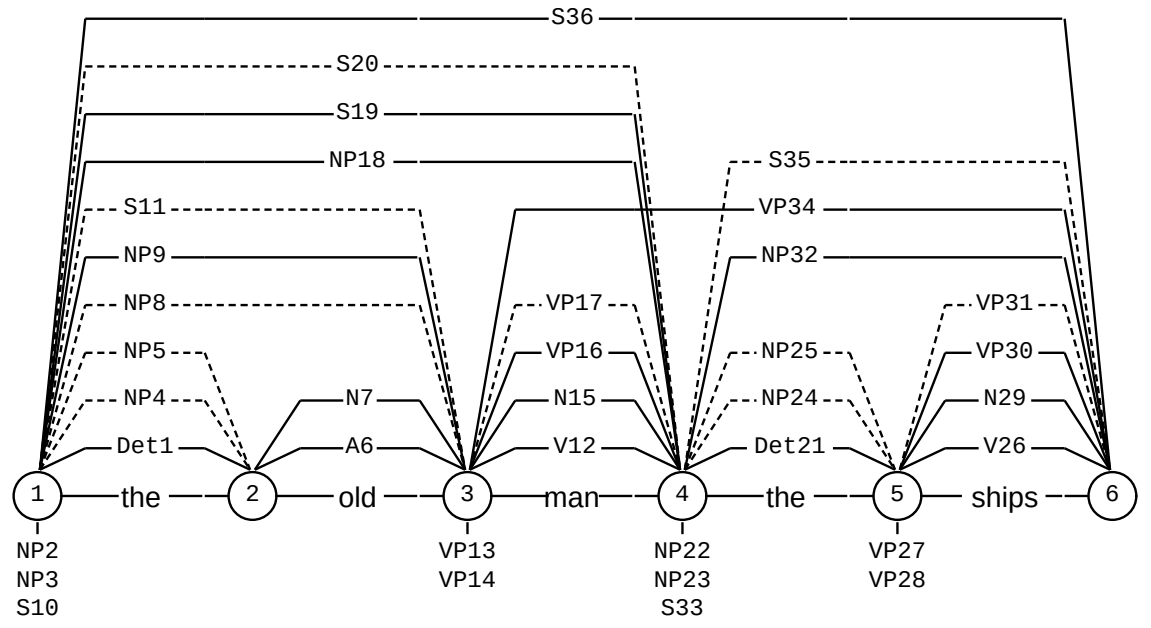

Figure 3: Chart of the sentence "The old man the ships" under bottom-up parsing.

mapping:

- Vertices $\tilde{v}_1, \ldots, \tilde{v}_{i-1} \in \tilde{V}$ correspond to $v_1, \ldots, v_{i-1} \in V$, respectively.
- Vertex $\tilde{v}_i$ corresponds to the "left half" of vertex $v_i$.
- Vertices $\tilde{v}_{i+1}, \ldots, \tilde{v}_{i+m-1} \in \tilde{V}$ do not correspond to any vertices in $V$.
- Vertex $\tilde{v}_{i+m}$ corresponds to the "right half" of vertex $v_i$.
- Vertices $\tilde{v}_{i+m+1}, \ldots, \tilde{v}_{n+1}$ correspond to $v_{i+1}, \ldots, v_{n+1-m}$, respectively.

The mapping is thus established with respect to insertion or deletion of an arbitrary number of contiguous tokens.[8]

## 3.3 Minimal Change

Assume that $E$ and $E'$ are the sets of edges of the charts $C$ and $C'$ obtained by batch-mode parsing of a text before and after a modification $\Delta_\tau$, respectively. We can then define the minimal output change on the basis of two edge sets as follows:

[8]Presumably, it is possible to generalize the mapping to more complex (non-contiguous) operations such as replacements or reversals. However, we do not pursue that here.

| | | | | | |
|---|---|---|---|---|---|
| $S$ | $\rightarrow$ | $NP\ VP$ | the | $\leftarrow$ | $Det$ |
| $NP$ | $\rightarrow$ | $Det\ N$ | old | $\leftarrow$ | $N,\ A$ |
| $NP$ | $\rightarrow$ | $Det\ A\ N$ | man | $\leftarrow$ | $N,\ V$ |
| $VP$ | $\rightarrow$ | $V$ | ships | $\leftarrow$ | $N,\ V$ |
| $VP$ | $\rightarrow$ | $V\ NP$ | | | |

Figure 4: Example grammar and lexicon.

**Definition 3 (Minimal output change)** We define the *set of missing edges* as the set difference $M = E \setminus E'$ and the *set of new edges* as the set difference $N = E' \setminus E$. We then define the *minimal output change* as $\Delta_{C_{min}} = M \cup N$.

Next, we can define the size of the minimal change as follows:

**Definition 4 (Size of minimal change)** We define the *size of the minimal change* as $\delta = |\Delta_\tau| + |\Delta_{C_{min}}|$, the sum of the number of inserted or deleted tokens and the number of edges in $\Delta_{C_{min}}$.

## 3.4 An Example

As an illustration, the chart in figure 2 is obtained under (batch-mode) bottom-up parsing, given the grammar in figure 4 and the sentence "The old man the tall ships". If the token "tall" is removed, the chart in figure 3 is obtained. Vertex $v_5$ in figure 2 then corresponds to the left half of vertex $v'_5$ in figure 3, and vertex $v_6$ corresponds to the right half of vertex $v'_5$. Furthermore, $v_7$ corresponds to $v'_6$. Clearly, the input change $\Delta_\tau$ consists of the token "tall". The output change $\Delta_{C_{min}}$ consists of the missing set $M$, which contains the three edges $A_{26}$, $NP_{27}$ and $NP_{34}$ in figure 2, and the new set $N$, which contains the single edge $NP_{32}$ in figure 3. The size of the change is then $\delta = |\Delta_\tau| + |\Delta_{C_{min}}| = 1 + 3 + 1 = 5$.

If instead "tall" is inserted before the last word in the sentence in figure 3, then the input change still consists of the token "tall". However, the two sets making up the output change are reversed: the missing set contains the single edge $NP_{32}$ in figure 3 and the new set contains the three edges $A_{26}$, $NP_{27}$ and $NP_{34}$ in figure 2. Thus, the size of the change is again 5.

# 4 An Unbounded Algorithm

A key idea of the incremental chart-parsing algorithm put forward by Wirén [14, 15] is to use edge dependencies for keeping track of edges that have to be removed in response to a change. An edge $e'$ is said to depend upon another edge or token $e$ if it is formed (derived) directly on the basis of $e$. Furthermore, if $e'$ is redundantly proposed by an edge $f$, then $e'$ can be said to depend (also) on $f$. By $e'$ being "redundantly proposed", we mean that the parser attempts to add an edge that is equivalent to $e'$ to the chart, but that that edge is rejected by the standard redundancy test in chart parsing. In effect, $f$ provides an additional "justification" for $e'$.

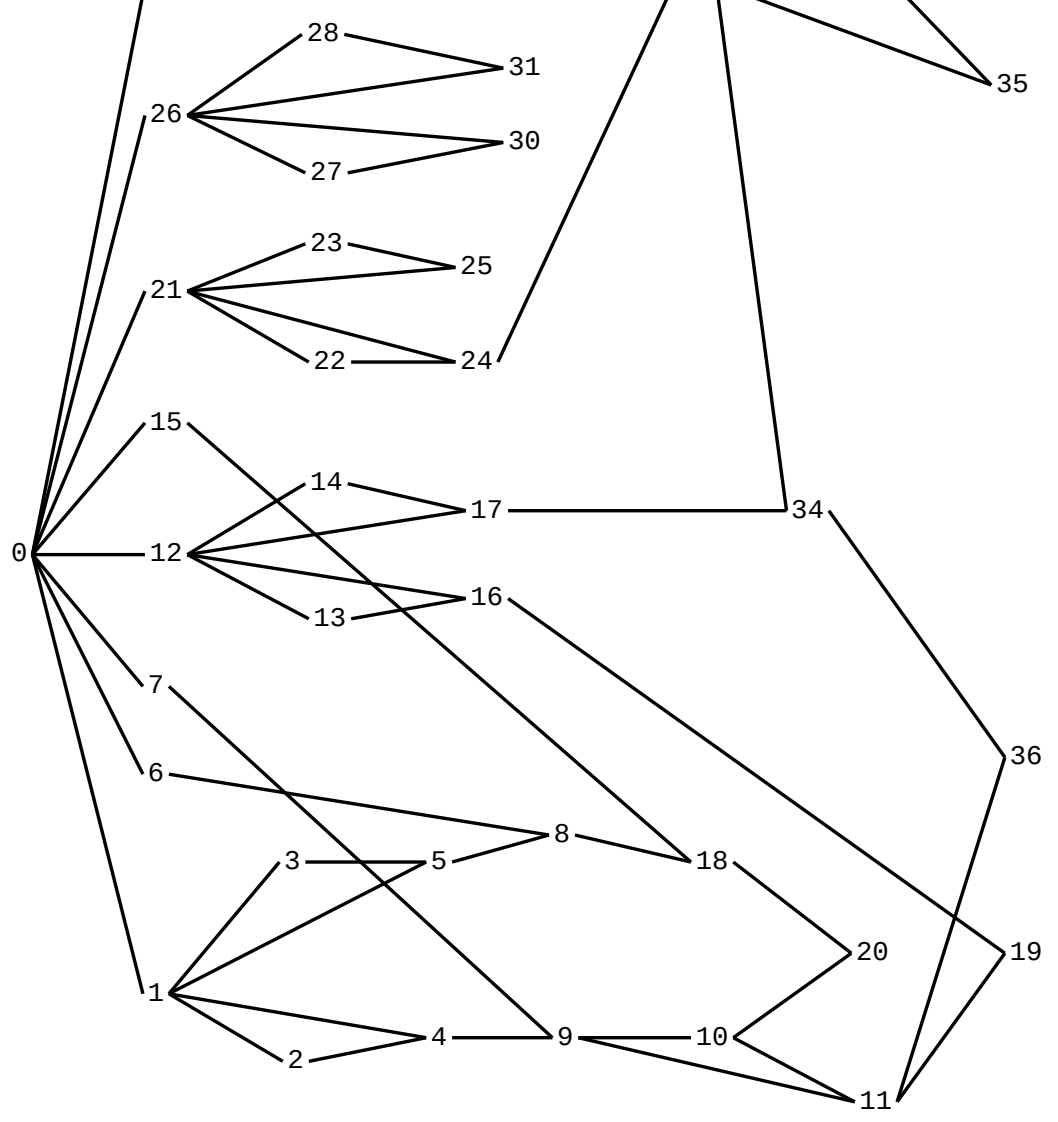

Figure 5: Edge-dependency graph induced by $\mathcal{D}$. The nodes of the graph correspond to the chart edges in figure 3. A dummy root node 0 is shown instead of nodes corresponding to the tokens.

Given a chart $C = \langle V, E \rangle$ and a set of tokens $\tau$, these conditions correspond to the following dependency relation on $E$ and $\tau$:

**Definition 5 (Edge dependency)** We define $\mathcal{D}$ as a binary relation on the set of chart edges and the set of tokens $E \cup \tau$ such that $\mathcal{D}(s, d)$ holds if and only if $d \in E$ is formed, or is redundantly proposed, directly using $s \in E \cup \tau$ according to a chart-parsing algorithm. We say that $d$ is a *dependent* (or *derivative*) edge of $s$, and that $s$ is a *source* edge (token) of $d$.

$\mathcal{D}$ can be illustrated by a graph. The dependency graph corresponding to the chart in figure 3 is shown in figure 5.

On the basis of the dependency relation, Wirén and Rönnquist [15, 17] define different disturbance sets, given as functions from tokens to sets of edges, and containing edges that need to be removed from the chart in response to a token-level change. The simplest such set is $\mathcal{D}^*(t_j)$, the transitive closure of $\mathcal{D}(t_j)$. Wirén and Rönnquist [15, 17] discuss this and other alternatives and show completeness of $\mathcal{D}^*$ with respect to the missing set.

The algorithm performs an update essentially by removing the entire disturbance set and then generating all possible edges. The latter set includes not only the new edges, but also disturbed, non-missing edges, which have to be generated anew. The complexity analysis of the algorithm yields that it is unbounded incremental in both its bottom-up and top-down version (see Wirén [16]). The source of this is that the algorithm removes the entire disturbance set, whose size depends on $n$, the size of the entire input.

# 5 A Bounded Algorithm

## 5.1 Intuitive Idea

Intuitively, a bounded incremental algorithm only processes the region where the input or output changes during an update cycle. In our case, the problem in achieving this is that the missing and new edges are not a priori known — when the incremental update begins, only a set of potentially missing edges (the disturbance set) is known. However, the update can be limited by using a change-propagation algorithm (compare Ramalingam and Reps [10, page 21]): By initially retaining the disturbance set, new and old edges can be compared during reparsing. If a new edge $e'$ is different from the corresponding old edge $e$ (if this exists), then the dependants of $e$ are regarded as disturbed (potentially missing). If $e'$ is equivalent to $e$ in the sense of giving rise to the same derivative edges, then the dependants of $e$ are known not to be missing, and hence the reparsing process does not have to proceed beyond this point in the search space. In order to avoid extra computation, the disturbed edges should be visited in the order given by the dependency graph.

How can the points at which a change "dies out" be characterized? Since we are interested in characterizing the conditions under which two edges give rise to the same derivative edges, the contents part of an edge (that is, the right-hand side before the dot of the dotted rule) is irrelevant. For example, we want to say that the new edge $\mathrm{NP}_{32}$ in figure 3 to be reparsing-equivalent with edge $\mathrm{NP}_{34}$ in figure 2 although their dotted rules and parse trees are different: the dotted rule of the former is $NP \rightarrow Det\ N\,\bullet$ and that of the latter is $NP \rightarrow Det\ A\ N\,\bullet$. We can summarize this in the following definition:

**Definition 6 (Reparsing-equivalent edges)** Assume given a proposed edge $e$ and a disturbed edge $e' \in C$. We say that $e = \langle v_i, v_j, X \rightarrow \alpha \bullet \beta \rangle$ and $e' = \langle v_s, v_t, Y \rightarrow \mu \bullet \nu \rangle$ are equivalent from the point of view of reparsing if $i = s$, $j = t$, $X = Y$ and $\beta = \nu$.

Inactive (combined or lexical) edges and predicted edges are special cases under this definition. In the former case, $\beta$ and $\nu$ are empty, and thus two inactive edges are reparsing-equivalent if $i = s$, $j = t$ and $X = Y$. In the latter case, $\alpha$ and $\mu$ are empty, and thus two predicted edges $e$ and $e'$ are reparsing-equivalent if $e = e'$.

## 5.2 The Algorithm

We now specify a bounded incremental chart-parsing algorithm that handles one update cycle.[9] In compari-son with the unbounded algorithm, the differences are in the reparse and remove steps.

[9] The algorithm is currently being implemented.

**Algorithm 2 (Incremental Chart Parsing)**

**Input:** A configuration $\langle \tau, C, \mathcal{D} \rangle$ and a modification $\Delta_\tau$ corresponding to insertion or deletion of $m$ tokens $t_i, \ldots, t_{i+m}$.

**Output:** An updated configuration $\langle \tau', C', \mathcal{D}' \rangle$.

**Method:** Do the following steps:

**Modify the problem instance:** Insert or delete the modified tokens given by $\Delta_\tau$ into or from $\tau$.

**Prepare the chart:** Do one of the following steps in the case of insertion or deletion, respectively:

*Insertion*: Renumber edges as follows: First, replace each edge $\langle v_j, v_k, r \rangle$ where $j \geq i$ and $k \neq i$ with an edge $\langle v_{j+m}, v_k, r \rangle$. Secondly, replace each edge $\langle v_j, v_k, r \rangle$ where $k > i$ with an edge $\langle v_j, v_{k+m}, r \rangle$. Looping edges at the "modification vertex", which have the form $\langle v_i, v_i, r \rangle$, are dealt with differently depending on where their sources are located, which in turn depends on the prediction strategy:

- Bottom-up case: If the looping edge depends on an outgoing, non-looping edge ($\langle v_i, v_j, r \rangle$ such that $j > i$), then the looping edge is replaced with an edge $\langle v_{i+m}, v_{i+m}, r \rangle$ (in effect, it is moved).
- Top-down case: If the looping edge depends on an incoming, possibly looping edge ($\langle v_j, v_k, r \rangle$ such that $k \leq i$), then do nothing.

Finally, update the dependency relation $\mathcal{D}$ so that any edge $\langle v_j, v_k, r \rangle$ such that $j \leq i$ and $k > i$ is made dependent on $t_i$.

*Deletion*: Renumber edges as follows: First, replace each edge $\langle v_j, v_k, r \rangle$ where $j > i$ with an edge $\langle v_{j-m}, v_k, r \rangle$. Then replace each edge $\langle v_j, v_k, r \rangle$ where $k > i$ with an edge $\langle v_j, v_{k-m}, r \rangle$.

**Reparse:** Do the following steps:

In the case of insertion: create a scanning task for each new token; create a combination task for each active–inactive edge pair meeting at $v_i$ and $v_{i+m}$.

In the case of deletion: create a combination task for each active–inactive edge pair meeting at $v_i$.

Reparse while visiting the disturbed edges in the order given by the dependency graph and treating the disturbed edges as "sleeping" (that is, they do not play any role in the parsing process as such). Whenever a new edge is proposed, check if an equivalent edge exists in the disturbance set according to definition 6. If so, install the new edge, update $\mathcal{D}$ by letting the new edge inherit the dependencies from the old edge. Do not add any agenda items for the new edge (thereby discontinuing reparsing along this path). Mark the new edge as re-created (with respect to a reparsing-equivalent one).

**Remove edges:** Remove each edge that is in the disturbance set but not in the dependency set of any re-created edge.

## 5.3 Incremental Complexity

For the purpose of analysing the incremental complexity of algorithm 2, we assume that adding or removing an edge takes unit time. We also assume that no edge has more than a constant number of sources or dependants and, hence, that the time required to install or examine the dependencies of $k$ edges is $O(k)$.[10]

We first focus on the reparsing step.[11] Consider the case of a deletion within a text. The set of new edges $N$ are generated as a result of joining two subcharts, which we assume have length $i$ and $j$ and contain $O(i^2)$ and $O(j^2)$ edges, respectively (disregarding the grammar constant $|G|$). The joined chart thus has length $i+j$ and consists of $O((i+j)^2)$ edges. The number of new edges resulting from joining the subcharts is then $|N| = O((i+j)^2) - (O(i^2) + O(j^2)) = O(i \cdot j)$ edges. Since the algorithm generates these edges by invoking a $O(n^3)$ reparsing algorithm, the new edges require $O((i+j)^3) - (O(i^3) + O(j^3)) = O(i \cdot j \cdot (i+j)) = O(i^2 \cdot j^2) = O(|N^2|)$ time. The insertion case can be obtained in a similar way and gives the same result. In the remove step, the missing edges are found by following dependency chains originating from tokens until a reparsing-equivalent edge is found or the chain ends. This step can therefore be executed in $O(|M|)$ time. The algorithm as a whole then requires $O(\delta^2)$ time.

# 6 Conclusions

The boundedness criterion used here provides a guarantee that the next update state is never more than an amount of computation away from the current state that is limited by the size of the change. This criterion is very strong. It can be thought of as constituting one extreme point of a continuum of ways in which to measure the complexity of incremental algorithms. At the other extreme, we have the option of using $|P'| + |S'|$, the cost of discarding the old solution and invoking a batch-mode algorithm on the modified problem instance. This measure might be used for showing that an algorithm with poor worst-case incremental behaviour is still practical: Poor incremental behaviour means that the algorithm does not respond quickly to (some) small changes. However, it may still perform better than discarding the old solution and invoking a batch-mode algorithm. In other words, even if the algorithm is unbounded in $\delta$, it may have a lower time bound in $|P'| + |S'|$ than the batch-mode algorithm. The unbounded algorithm described in section 4 is an example of this: it is clearly more efficient than the batch-mode algorithm for the purpose of incremental update.

[10] This assumption is considered too strong in reason maintenance, where, in the worst case, all formulas can be directly connected (see Goodwin [4, page 110 f.]). However, it seems appropriate here, since under a context-free grammar of the kind used here only predicted edges may have multiple sources. Moreover, the number of these sources is limited because of the linearity of the problem instance (the text).

[11] Since we take addition and removal of edges to be the primary tasks of incremental update, we disregard the chart-preparation step. Although a more thorough analysis might take this step into account both in the definition of $\delta$ and in the complexity analysis, we do not believe that anything fundamental would be altered by this.

Several interesting topics for further research present themselves: One is to generalize the notions of minimal change and bounded incrementality to other processing frameworks that make use of a table or a chart, for example, pseudo-parallel LR parsing (Tomita [13]) or tabular generation (Neumann [8]). Another interesting topic is to translate the same notions to a unification-based grammar formalism. Defining minimal change then requires a definition of the difference between two feature structures. An immediate observation is that this is itself hardly a feature structure, but rather the set of (sub)structures that are not present in both feature structures (in analogy with our definition of the difference between two charts).